\pdfoutput=1
\PassOptionsToPackage{sort}{natbib}
\documentclass[11pt]{article}

\usepackage[final]{acl}

\usepackage{times}
\usepackage{latexsym}

\usepackage[T1]{fontenc}

\usepackage[utf8]{inputenc}

\usepackage{microtype}

\usepackage{inconsolata}

\usepackage{import}
\usepackage{graphicx}
\usepackage{amsmath}
\usepackage{amssymb}
\usepackage{booktabs}
\usepackage{float}
\usepackage{listings}
\usepackage{multirow}
\usepackage{lipsum}  
\usepackage{subcaption}
\usepackage{comment}
\usepackage[switch]{lineno}
\usepackage{wrapfig}
\usepackage{varwidth}
\usepackage{array}
\usepackage{makecell}
\usepackage[capitalize]{cleveref}

\usepackage[normalem]{ulem}
\useunder{\uline}{\ul}{}

\usepackage{pifont}

\newcommand{\rinlinecode}[1]{\lstinline[basicstyle=\ttfamily, breaklines=true, breakatwhitespace=true]|#1|}

\title{If CLIP Could Talk: Understanding Vision-Language Model Representations Through Their Preferred Concept Descriptions}

\author{
\textbf{Reza Esfandiarpoor}$^{1}$ \quad
\textbf{Cristina Menghini}$^{2}$ \quad
\textbf{Stephen H. Bach}$^{1}$
\\
$^{1}$Department of Computer Science, Brown University \\
$^{2}$Data Science Institute, Brown University \\
\texttt{\{reza\_esfandiarpoor,cristina\_menghini,stephen\_bach\}@brown.edu}
}

\begin{document}
\maketitle
\begin{abstract}

Recent works often assume that Vision-Language Model (VLM) representations are based on visual attributes like shape.
However, it is unclear to what extent VLMs prioritize this information to represent concepts.
We propose Extract and Explore (EX2), a novel approach to characterize textual features that are important for VLMs.
EX2 uses reinforcement learning to align a large language model with VLM preferences and generates descriptions that incorporate features that are important for the VLM.
Then, we inspect the descriptions to identify features that contribute to VLM representations.
Using EX2, we find that spurious descriptions have a major role in VLM representations despite providing no helpful information, e.g., \rinlinecode{Click to enlarge photo of CONCEPT}.
More importantly, among informative descriptions, VLMs rely significantly on non-visual attributes like habitat (e.g., North America) to represent visual concepts.
Also, our analysis reveals that different VLMs prioritize different attributes in their representations.
Overall, we show that VLMs do not simply match images to scene descriptions and that non-visual or even spurious descriptions significantly influence their representations.
Code: \url{https://github.com/BatsResearch/ex2}
\end{abstract}

\section{Introduction}
\label{sec:intro}

The ability of contrastive Vision-Language Models (VLMs) to match related text and images depends significantly on the content of text descriptions~\citep{novack2023chils,radford2021learning}.
Heuristically, to better identify the related images, humans seek more detailed information about the physical appearance of concepts. 
Many recent works extend this heuristic to VLMs and augment the descriptions (also called prompts) with additional visual information (e.g., shape) to more accurately identify the related images~\citep{menon2023visual,yan2023learning}.
However, in practice, even augmenting the descriptions with random words improves the performance~\citep{roth2023waffling}, and it is unclear what factors contribute to VLM concept representations.
Here, we propose a novel analysis method for characterizing textual features that contribute to the VLM representation of a concept.
Simply put, our goal is to understand how VLMs encode different concepts.

\begin{figure}[t!]
  \centering
  \includegraphics[width=\columnwidth]{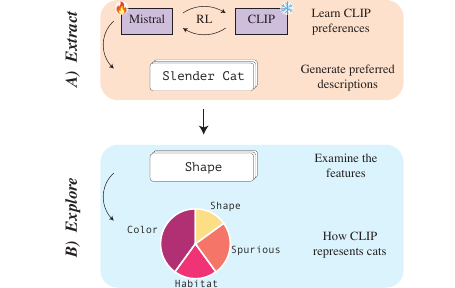}
  \caption{Extract: we align Mistral with VLM preferences and generate descriptions that contain features that are important for the VLM. Explore: we examine various aspects of these descriptions to identify features that contribute to VLM representations.}
  \label{fig:intro_fig}
\end{figure}

Despite the growing body of work on various aspects of VLMs (e.g., sensitivity to word order)~\citep{yuksekgonul2022and,Ma_2023_CVPR,akula-etal-2020-words}, there is limited understanding of VLMs' world knowledge acquired during pre-training.
\citet{yun2021does} show that VLMs and Large Language Models (LLMs) perform similarly on language tasks that require knowledge of the physical world.
\emph{If VLMs do not learn about the physical world from images, what do they learn?}
Moreover, \citet{yun2022vision} show that, the VLM representation of a concept is often not based on its visual attributes like color and shape.
\emph{If not based on visual attributes, how do VLMs represent concepts?}
Our work introduces a new analysis method for studying such exploratory questions.
Unlike previous works that focus on a very specific question, our goal is to explore and characterize the features that contribute to VLM representations with the feedback from the VLM itself.

In this paper, we propose Extract and Explore (EX2), a novel analysis method for understanding how VLMs represent concepts.
Instead of directly probing the VLM, EX2 uses reinforcement learning (RL) to align an LLM with VLM preferences.
Our reward function measures the similarity between the concept description and its images.
Thus, it encourages the LLM to generate descriptions that incorporate features that are important for the VLM.
Then, we can inspect these descriptions from various perspectives to identify common factors that contribute to VLM representations.
In this work, we examine the descriptions to understand how often VLMs prioritize additional information to represent concepts and how often this information explains the physical appearance of concepts.

We use EX2 to analyze seven different VLMs on six fine-grained classification datasets.
The aligned LLM successfully learns features that are important for the VLM and generates descriptions that achieve better classification accuracy than a set of generic descriptions.
We find that spurious descriptions (e.g., \rinlinecode{Click to enlarge photo of CONCEPT}) have a major role in VLM representations despite providing no helpful information about concepts.
More importantly, even when VLMs prioritize informative descriptions, they significantly rely on non-visual attributes\footnote{Throughout the paper, ``attribute'' refers to concept attributes like color and shape, while ``feature'' and ``characteristic'' refer to properties of descriptions like the type of information they contain or text style.} like habitat (e.g., North America) to represent visual concepts.
In our fine-grained analysis, we find that different VLMs represent similar concepts differently, i.e., prioritize different attributes.
Even the same VLM prioritizes different attributes for different datasets, suggesting that extensive experiments across different datasets are required to draw conclusions about VLMs.
EX2 is specifically suited to automatically carry out such analysis on a wide range of existing classification datasets.
Finally, we show EX2's application for hypothesis generation and suggest new research questions based on our findings.

Our findings encourage further work on image-text pre-training to address VLMs' reliance on spurious descriptions and the alignment between non-visual information and images.
Moreover, EX2 provides the future work with a flexible tool to study the impact of different pre-training methods and datasets from various aspects.
We summarize our contributions and findings as following:

\begin{itemize}
    \item We propose EX2, a novel analysis method that aligns an LLM with VLM preferences in order to characterize textual features that contribute to VLM representations.

    \item We show that EX2 successfully learns features that are important for the VLM and generates descriptions that improve the downstream classification accuracy, highlighting the benefits of adapting to VLM preferences for downstream tasks as well as analysis.

    \item We show that VLMs significantly rely on spurious or non-visual descriptions to represent visual concepts. We find that different VLMs prioritize different attributes to represent similar concepts. Even the same VLM prioritizes different attributes across datasets, emphasizing the benefits of EX2's ability to automatically analyze VLMs on a wide range of existing classification datasets.
\end{itemize}

\begin{figure*}[t!]
  \centering
  \includegraphics[width=\textwidth]{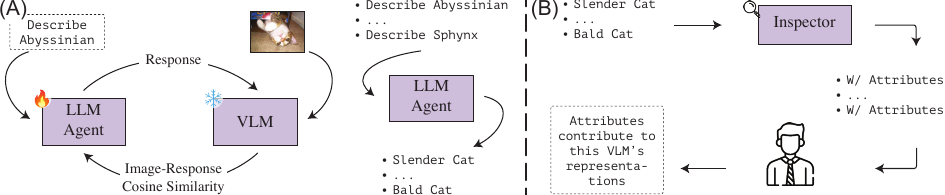}
  \caption{Extract and Explore (EX2) overview. A) We use RL to fine-tune an LLM to generate concept descriptions that are closer to the corresponding images in the VLM embedding space, thus, the descriptions incorporate features that the VLM uses to represent the concepts. We use the aligned LLM to generate the VLM's preferred description for all concepts. B) We inspect these descriptions from various aspects, e.g., if they are informative or describe visual attributes. Based on the aggregate results, we draw conclusions about how the VLM represents concepts.}
  \label{fig:block_diagram}
\end{figure*}

\section{Related Work}
\label{sec:related_work}

\textbf{Reinforcement Learning for Language Models}\hphantom{A}
In recent years, RL has been successfully used to align LLMs with human preferences, known as reinforcement learning with human feedback (RLHF)~\citep{ramamurthy2023is,bai2022training}.
Instead of human preferences, we use similar methods to~\citet{stiennon2020learning} and~\citet{ziegler2019fine} to align the LLM with VLM's preferences for descriptions that it deems more accurate.
Unlike previous methods, we use preference learning not as our goal but as a tool to understand the knowledge that VLMs acquire during pre-training.

\noindent\textbf{Vision-Language Models (VLMs)}\hphantom{A}
We primarily focus on contrastive VLMs, which are trained to push related image-text pairs closer and unrelated pairs farther in the embedding space~\citep{radford2021learning,jia2021scaling}.
Multimodal language models (MLMs) are another category of VLMs that use additional training to incorporate the image features into LLMs in order to condition the text generation on both input text and images~\citep{li2023blip,liu2023llava}.
Despite the many advantages of MLMs, contrastive VLMs are still better suited for tasks like image classification~\citep{alayrac2022flamingo} or creating large search indices for efficient filtering and retrieval~\citep{schuhmann2022laion}.
Contrastive VLMs also serve as the backbone for many other methods or downstream applications like image manipulation~\citep{patashnik2021styleclip}.
Even most MLMs use contrastive VLMs to extract the image features~\citep{liu2023llava}.
Thus, understanding how contrastive VLMs represent concepts remains an important issue that also impacts other types of VLMs and many downstream applications.
We aim to investigate this question without fine-tuning or modifying the VLM after pre-training.

\noindent\textbf{Detailed Descriptions for VLMs}\hphantom{A}
One line of work provides VLMs with detailed concept descriptions to improve the classification accuracy~\citep{yang2023language,pratt2023does,yan2023learning,li2023desco,feng2023text}.
For instance,~\citet{menon2023visual} suggest augmenting the descriptions with concept attributes.
\citet{esfandiarpoor2024followup} further improve this idea and ensure the attributes differentiate the target classes.
However,~\citet{roth2023waffling} question the role of additional information and suggest that even adding random words and characters to descriptions leads to similar improvements.
Here, we propose a novel model analysis method to study the role of additional information in VLM representations.

\noindent\textbf{Vision-Language Model Analysis}\hphantom{A}
The massive success of VLMs has sparked an interest in understanding how they interpret text descriptions.
Previous works have studied VLMs from various aspects such as the relative importance of verbs and nouns~\citep{hendricks-nematzadeh-2021-probing}, VLMs' sensitivity to word order~\citep{akula-etal-2020-words}, linguistic features of descriptions~\citep{castro-etal-2023-scalable}, and, the most popular, their poor compositional capabilities~\citep{schiappa2023probing,xu2024benchmarking,lewis2022does,hsieh2023sugarcrepe,thrush2022winoground,parcalabescu-etal-2022-valse,Ma_2023_CVPR}.
Although this line of work provides important information about specific aspects of VLMs' behavior, our understanding of the world knowledge that VLMs acquire during pre-training remains limited.
As discussed in~\cref{sec:intro}, this becomes even more important considering that previous works suggest VLMs' behavior deviates from the expected results of vision and language pre-training~\citep{yun2021does,yun2022vision}.
Given the importance of concept representations~\citep{merullo2022linearly,patel2021mapping,lovering2022unit,pavlick2022semantic},
our goal is to understand what VLMs learn during pre-training and characterize factors contributing to their concept representations.

Moreover, we propose a new paradigm for studying VLMs that allows us to conduct such exploratory analysis.
Currently, the most common approach for VLM analysis relies on custom datasets that test a specific hypothesis.
However, our approach is compatible with existing classification datasets and is not tied to a specific hypothesis.
We can even use it to generate new hypotheses (\cref{sec:hypothesis_gen}).
There is also a complementary line of work for understanding VLM representations in terms of images instead of text~\citep{kazemi2024we,ghiasi2022vision}.
For natural language tasks,~\citet{perez2022red} use a similar method, i.e., LLM alignment, to discover the harmful generations of other LLMs.
We emphasize that our method is not meant to replace existing VLM analysis approaches but to provide a complementary tool that allows for exploratory model analysis.

\section{Extract and Explore}
\label{sec:method}

Given the large number of textual features that could contribute to VLM representations (e.g., concept attributes and linguistic patterns), it is very costly, if not impossible, to curate and probe VLMs for an exhaustive set of features.
Instead, EX2 learns to automatically generate concept descriptions that incorporate textual features that are important for the VLM, i.e., Extract VLM's preferred descriptions (\cref{fig:block_diagram}a).
Then, EX2 inspects these descriptions for common patterns to identify features that the VLM uses to represent concepts, i.e., Explore VLM's preferred descriptions (\cref{fig:block_diagram}b).

\subsection{Extracting VLM's Preferred Descriptions}

\textbf{A Large Search Space}\hphantom{i}
To expand the search space for potential features, we use LLMs to generate concept descriptions.
LLM's ability to control various aspects of text generation, such as world knowledge and text style~\citep{petroni2019language,jiang2020can}, leads to a more thorough search than a predefined set of features.
To further expand the search space, we use 25 different questions to query the LLM about various aspects of each concept (refer to~\cref{sec:imp_details} for the list of question templates).

\textbf{LLM Alignment with VLM Preferences}\hphantom{A}
To increase the likelihood of VLM's desired features in descriptions, we use reinforcement learning to align the LLM with VLM preferences.
Thus, we define a reward function that gives a higher score to descriptions that include features that are important for the VLM.
Intuitively, from the VLM's perspective, the reward function encourages more accurate concept descriptions.
Since contrastive VLMs like CLIP are trained to push related text and images closer, VLMs deem a concept description accurate if it is close to the embedding of the corresponding images.
Therefore, we define the reward function as the average cosine similarity between the concept description and concept images.

Specifically, for some description $d_c$ and a set of images $D_c$ for class $c$, we calculate the reward as:
\begin{equation*}
    R(d_c) = \frac{1}{|D_c|} \sum_{x \in D_c} \tau\cos(\Phi_I(x), \Phi_T(d_c)) - \beta KL \, ,
\end{equation*}
where $\Phi_I$ and $\Phi_T$ are the VLM image and text encoders, respectively, $\tau$ is a constant scaling factor, and $\beta$ is the coefficient for the KL divergence between the original and current model.
The KL penalty encourages the generation of meaningful descriptions that we can later analyze and helps with convergence by constraining the search space.
Since we want to compare the generations across experiments, we adopt the adaptive KL coefficient technique of~\citet{ziegler2019fine} to achieve roughly similar KL divergences across experiments.

With the 25 questions for all classes as the LLM input and this reward function, we use the same method as~\citet{stiennon2020learning} and~\citet{ziegler2019fine}, which uses Proximal Policy Optimization~\citep{schulman2017proximal}, to update the LLM to increase the reward score.

Then, for each concept, EX2 uses the aligned LLM to generate a set of descriptions that are similar to the corresponding images, i.e., they are the VLM's preferred descriptions for the concept.

\subsection{Exploring VLM's Preferred Descriptions}
\label{sec:explore}
Since the aligned descriptions incorporate features that are important for the VLM, we inspect these descriptions for desired characteristics to understand the extent of their contribution to VLM representations.
For instance, we can check if VLMs rely on attributes to represent concepts or if VLMs are biased towards a specific text style.
We use ChatGPT \citep{brown2020language,ouyang2022training} to automatically inspect the large number of descriptions.

\begin{figure}[h]
  \centering
  \includegraphics[width=\columnwidth]{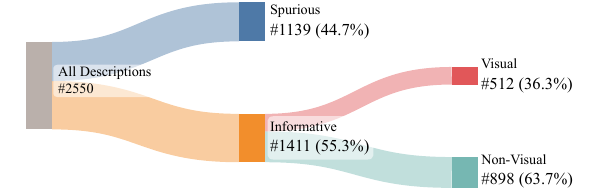}
  \caption{Breakdown of aligned descriptions for CLIP on Flowers. CLIP significantly relies on spurious or non-visual information to represent flower species.}
  \label{fig:sankey_diagram}
\end{figure}

\begin{table*}[t]
\centering
\setlength{\tabcolsep}{4pt}
\resizebox{\textwidth}{!}{
\begin{tabular}{@{}lcccccccccccc@{}}
\toprule

VLM
& \multicolumn{2}{c}{Flowers}     
& \multicolumn{2}{c}{Pets}   
& \multicolumn{2}{c}{CUB}         
& \multicolumn{2}{c}{Stanford Dogs} 
& \multicolumn{2}{c}{Aircrafts}   
& \multicolumn{2}{c}{Stanford Cars} \\
\midrule
                                     
& Temp.
& EX2
& Temp.
& EX2            
& Temp.
& EX2            
& Temp.
& EX2               
& Temp.
& EX2            
& Temp.
& EX2
\\
\cmidrule(lr){2-3}
\cmidrule(lr){4-5}
\cmidrule(lr){6-7}
\cmidrule(lr){8-9}
\cmidrule(lr){10-11}
\cmidrule(lr){12-13}

CLIP         
& 63.31          
& \textbf{73.65} 
& 84.68     
& \textbf{88.34} 
& 51.48          
& \textbf{53.54} 
& 59.59         
& \textbf{61.49}    
& 21.36          
& \textbf{22.89} 
& 60.15           
& \textbf{60.88}  \\
ALIGN                
& 59.44          
& \textbf{62.19} 
& 82.56     
& \textbf{86.86} 
& 36.81          
& \textbf{39.66} 
& 52.38         
& \textbf{56.24}    
& 12.42          
& \textbf{17.04} 
& 72.73           
& \textbf{74.15}  \\
EVA     
& 74.47          
& \textbf{79.05} 
& 91.80     
& \textbf{94.52} 
& \textbf{74.63} 
& 73.52          
& 76.08         
& \textbf{79.37}    
& 37.08          
& \textbf{39.66} 
& \textbf{91.62}  
& 91.22           \\
SigLIP                
& 82.48          
& \textbf{86.81} 
& 92.10     
& \textbf{94.49} 
& 61.96          
& \textbf{65.33} 
& 76.76         
& \textbf{78.89}    
& 48.21          
& \textbf{50.56} 
& 90.92           
& \textbf{91.22}  \\
MetaCLIP    
& 71.44          
& \textbf{73.18} 
& 87.84     
& \textbf{91.42} 
& 69.88          
& \textbf{69.99} 
& 65.50         
& \textbf{68.87}    
& \textbf{37.50} 
& 36.72          
& \textbf{76.22}  
& 76.17           \\
CLIPA           
& 77.13          
& \textbf{79.61} 
& 91.93     
& \textbf{93.87} 
& 74.56          
& \textbf{76.99} 
& 77.19         
& \textbf{79.34}    
& 39.75          
& \textbf{41.31} 
& \textbf{94.80}  
& \textbf{94.76}  \\
DFN 
& \textbf{88.96} 
& 88.86          
& 94.33     
& 94.33          
& 87.25          
& \textbf{87.78} 
& 84.31         
& \textbf{85.75}    
& \textbf{78.76} 
& 75.88          
& \textbf{96.14}  
& 95.78           \\ \bottomrule
\end{tabular}
}
\caption{Accuracy of EX2 descriptions. Temp. is the set of 80 generic templates~\citep{radford2021learning}.}
\label{tab:main_classification}
\end{table*}

Here, we study VLMs as visual perception tools that are expected to match the description of a concept's physical appearance to images of instances of that concept.
Similar to the reward function, our analysis focuses on the physical appearance of concepts rather than the entire content of individual images.
For many applications, it is undesirable for models to rely on \emph{spurious} or \emph{non-visual} information. Also see~\cref{sec:discussion_rationale,sec:discussion_concept_level} for extended discussion.
We consider descriptions that provide no world knowledge about the concept to be spurious (e.g., \rinlinecode{click to enlarge photo of the cat}).
We consider informative descriptions to be non-visual if they only contain additional information that is not about the concept's physical appearance (e.g., \rinlinecode{The bobolink is a North American bird}).
For example, we find that CLIP relies significantly on both spurious and non-visual information to represent flowers, with 45\% of descriptions providing no helpful information and 64\% of informative descriptions explaining non-visual attributes (\cref{fig:sankey_diagram}).

The space of potential hypotheses goes beyond our specific analysis, and we encourage future work to study other aspects of the aligned descriptions, such as text style or linguistic patterns.
Furthermore, as we show in~\cref{sec:hypothesis_gen}, the trends in EX2 descriptions provide helpful cues for discovering new research questions for further investigation.

\section{Experiments}
\label{sec:experiments}

In this section, we use EX2 to analyze different VLMs at three levels (\cref{fig:sankey_diagram}).
We first validate that the LLM learns features that are important for each VLM, and aligned descriptions improve the classification accuracy.
Second, we show that spurious descriptions contribute significantly to VLM representations despite providing no helpful information.
More importantly, at the third level, we find that VLMs rely significantly on non-visual attributes to represent visual concepts.
We discuss potential correlations between pre-training details and our findings in~\cref{sec:discussion_findings_cause}.
We conduct a more fine-grained analysis that reveals even for the same dataset, different VLMs prioritize different attributes to represent the concepts.
Moreover, even the same VLM prioritizes different attributes for different datasets.
Finally, we show EX2's application for hypothesis generation and suggest new research questions based on our observations.

\subsection{Setup}

After RL fine-tuning, we query the LLM with the same questions used during fine-tuning (\cref{sec:method}) and generate 25 descriptions for each concept that incorporate features that are important for the VLM.
To study VLM preferences, our analysis is only based on a subset of generated descriptions that help VLMs better identify the related images and hence are representative of VLM preferences.
We consider a set of descriptions helpful if they achieve better image classification accuracy on the corresponding task than the ensemble of 80 generic descriptions used by~\citet{radford2021learning}.
Similar to~\citet{radford2021learning}, we use the 25 generated descriptions for each concept as a prompt ensemble and use cosine similarity to predict the label for each image.

\noindent\textbf{Datasets}\hphantom{A}
We use six classification datasets for analysis.
CUB200-2011: fine-grained bird species recognition~\citep{wah2011caltech}.
FGVCAircraft: aircraft model classification~\citep{maji13finegrained}.
Flowers102: fine-grained flower species recognition~\citep{Nilsback08}.
Oxford IIIT Pets~\citep{parkhi12a}.
Stanford Dogs~\citep{KhoslaYaoJayadevaprakashFeiFei_FGVC2011}.
Stanford Cars~\citep{krause20133d}.

\noindent\textbf{Models}\hphantom{A}
We use Mistral-7B as our LLM~\citep{jiang2023mistral}.
We choose seven different VLMs for our analysis.
CLIP~\citep{radford2021learning}, ALIGN~\citep{jia2021scaling}, EVA~\citep{fang2023eva}, SigLIP~\citep{zhai2023sigmoid}, MetaCLIP~\citep{xu2023demystifying}, CLIPA~\citep{li2023inverse}, and DFN~\citep{fang2023data}.
Refer to~\cref{sec:models} for details.

\begin{table*}[t]
\centering
\begin{tabular}{@{}lcccccc@{}}
\toprule

& Flowers        
& Pets           
& CUB            
& Stanford Dogs  
& Aircrafts      
& Stanford Cars  
\\
\cmidrule(l){2-7}

CLIP   
& \textbf{55.33} 
& \textbf{61.73} 
& \textbf{83.50} 
& \textbf{49.03} 
& \textbf{57.84} 
& \textbf{44.47} 
\\
ALIGN    
& \textbf{63.92} 
& \textbf{68.76} 
& \textbf{63.66} 
& \textbf{73.87} 
& \textbf{65.60} 
& \textbf{42.88} 
\\
EVA      
& 1.80           
& 0.00           
& -
& \textbf{36.60} 
& \textbf{29.20} 
& -
\\
SigLIP   
& \textbf{44.59} 
& 18.70          
& 23.06          
& 21.00          
& 5.36           
& 9.96           
\\
MetaCLIP 
& \textbf{43.06} 
& \textbf{32.22} 
& \textbf{26.58} 
& \textbf{51.40} 
& -
& -
\\
CLIPA    
& \textbf{55.18} 
& \textbf{26.70} 
& \textbf{47.84} 
& \textbf{78.80} 
& 6.24           
& -
\\
DFN      
& -
& -
& 0.32           
& 22.90          
& -
& -
\\

\bottomrule
\end{tabular}
\caption{The percentage of informative descriptions for experiments that the LLM successfully learns the VLM preferences and improves the classification accuracy. Bold numbers are $>$ 25\%.}
\label{tab:has_info}
\end{table*}

\subsection{Successful Alignment}

We use classification accuracy to verify that the LLM learns what features contribute to VLM representations.
In 33 out of 42 experiments, the LLM successfully learns what features help the VLM to identify the related images, and the aligned descriptions improve the accuracy compared to the generic template set (\cref{tab:main_classification}).
Notably, most of the other nine experiments involve the Stanford Cars dataset or the DFN model.
Although our goal is model analysis, our results also emphasize the benefits of adapting to VLM preferences for downstream tasks like classification.
In the remainder of this section, we analyze the descriptions for the 33 experiments where we successfully learn the VLM preferences to characterize textual features that contribute to concept representations.

\subsection{Informative vs Spurious Descriptions}

Heuristically, we expect descriptions to better represent concepts when they contain additional information and make little to no difference otherwise.
To study the role of spurious descriptions in VLM representations, we inspect if descriptions provide additional information about concepts.

\begin{table}[h!]
\centering
\setlength{\tabcolsep}{4pt}
\resizebox{\columnwidth}{!}{
\begin{tabular}{@{}llccc@{}}
\toprule

Dataset       
& VLM      
& Temp. Set      
& w/o Info       
& w/ Info
\\
\midrule
              
& CLIP   
& 63.31          
& {\ul 72.19}    
& \textbf{73.77} \\
              
& ALIGN    
& {\ul 59.44}    
& 59.39          
& \textbf{61.90} \\

Flowers       
& SigLIP   
& 82.48          
& {\ul 85.43}    
& \textbf{87.05} \\
              
& MetaCLIP 
& {\ul 71.44}    
& \textbf{73.44} 
& 70.39          \\
              
& CLIPA    
& 77.13          
& {\ul 79.36}    
& \textbf{79.40} \\
\midrule

& CLIP   
& 84.68          
& {\ul 87.49}    
& \textbf{87.90} \\
Pets          
& ALIGN    
& 82.56          
& {\ul 85.61}    
& \textbf{86.54} \\
              
& MetaCLIP 
& 87.84          
& \textbf{91.61} 
& {\ul 90.22}    \\
              
& CLIPA    
& 91.93          
& \textbf{94.09} 
& {\ul 92.75}    \\
\midrule
              
& CLIP   
& {\ul 51.48}    
& 46.50          
& \textbf{52.88} \\
CUB           
& ALIGN    
& 36.81          
& {\ul 36.99}    
& \textbf{39.47} \\
              
& MetaCLIP 
& \textbf{69.88} 
& {\ul 69.43}    
& 67.88          \\
              
& CLIPA    
& 74.56          
& {\ul 75.94}    
& \textbf{76.73} \\
\midrule
              
& CLIP   
& 59.59          
& \textbf{60.71} 
& {\ul 60.34}    \\
              
& ALIGN    
& {\ul 52.38}    
& 52.32          
& \textbf{55.85} \\

Stanford
& EVA      
& 76.08          
& \textbf{79.41} 
& {\ul 78.07}    \\

Dogs              
& MetaCLIP 
& 65.50          
& \textbf{68.54} 
& {\ul 66.43}    \\
              
& CLIPA    
& {\ul 77.19}    
& 76.82          
& \textbf{79.09} \\
\midrule
              
& CLIP   
& 21.36          
& \textbf{23.58} 
& {\ul 22.17}    \\
Aircrafts     
& ALIGN    
& 12.42          
& {\ul 14.79}    
& \textbf{18.09} \\
              
& EVA      
& {\ul 37.08}    
& \textbf{39.09} 
& 36.63          \\
\midrule
              
Stanford
& CLIP   
& {\ul 60.15}    
& \textbf{60.17} 
& 59.74          \\

Cars 
& ALIGN    
& 72.73          
& {\ul 73.54}    
& \textbf{74.00} \\
              
\bottomrule
\end{tabular}
}
\caption{Classification accuracy of informative (w/ Info) and spurious (w/o Info) descriptions. First and second best numbers are in bold and underline.}
\label{tab:info_classification}
\end{table}

\textbf{Spurious descriptions are a major factor in VLM representations.}
\Cref{tab:has_info} reports the percentage of descriptions that provide additional information about concepts for the 33 cases in which the LLM successfully learns the VLM preferences.
We refer to descriptions that provide no additional information about concepts as spurious (e.g., \rinlinecode{Photo of CONCEPT attracted my attention}).
We observe that in 10 cases, the improvements are almost solely driven by spurious descriptions, and there is no considerable amount of additional information in descriptions, i.e., <25\% of descriptions are informative.
In a few of these cases, the LLM learns to only generate the concept names with some artifacts, without any helpful information (see~\cref{sec:qual_observ} for examples).
Notably, SigLIP benefits more from spurious descriptions (5/6 datasets) than other VLMs.
To understand the role of spurious descriptions in the remaining 23 cases, we separate the informative and non-informative descriptions and measure the classification accuracy (\cref{tab:info_classification}).
Fortunately, VLMs do not solely rely on spurious descriptions, and in 19 cases, the informative descriptions alone improve the performance.
However, in 16 cases, spurious descriptions alone lead to a considerable boost in accuracy.
In total, in 26 out of the 33 experiments, spurious descriptions are either almost solely responsible for the improvements or have a considerable contribution.

\subsection{Visual vs Non-visual Information}

VLMs are often thought to match images to scene descriptions.
Therefore, it is natural to think of descriptions of visual attributes, like color, as the basis for VLM representations and ignore or assume a negligible role for non-visual attributes like habitat (e.g., North America).
To investigate the contribution of non-visual attributes to VLM representations, we inspect if each informative description provides visual or non-visual information.

\begin{table}[h]
\centering
\begin{tabular}{@{}lccc@{}}
\toprule

VLM       
& Flowers
& Pets
& Stanford Dogs
\\
\midrule

CLIP
& \textbf{36.29} 
& \textbf{54.82} 
& \textbf{33.45} 
\\

ALIGN
& \textbf{56.44} 
& \textbf{55.66} 
& \textbf{55.55} 
\\

EVA
& -              
& -              
& \textbf{30.78} 
\\

SigLIP
& \textbf{67.46} 
& -              
& -              
\\

MetaCLIP
& -              
& \textbf{57.38} 
& \textbf{45.40} 
\\

CLIPA
& \textbf{58.71} 
& \textbf{64.78} 
& \textbf{60.03} 
\\

\toprule
\bottomrule

& CUB
& Aircrafts
& Stanford Cars
\\
\cmidrule(lr){2-4}

CLIP
& \textbf{64.00} 
& 10.24     
& -              
\\
ALIGN
& 18.69          
& 9.82      
& 8.66           
\\

CLIPA
& \textbf{39.34} 
& -         
& -              
\\

\bottomrule
\end{tabular}
\caption{Percentage of informative descriptions that contain visual attributes. Bold numbers are $>$ 25\%.}
\label{tab:is_visual}
\end{table}

\begin{table}[h]
\centering
\setlength{\tabcolsep}{4pt}
\resizebox{\columnwidth}{!}{
\begin{tabular}{@{}llccc@{}}
\toprule
Dataset             
& VLM       
& Temp. Set            
& Vis            
& Non-Vis        \\
\midrule
                    
& CLIP    
& 63.31                
& {\ul 66.92}    
& \textbf{73.54} \\
Flowers             
& ALIGN     
& 59.44                
& {\ul 60.61}    
& \textbf{62.34} \\
                    
& SigLip    
& 82.48                
& \textbf{86.84} 
& {\ul 85.10}    \\
                    
& CLIPA     
& 77.13                
& \textbf{79.79} 
& {\ul 79.22}    \\
                    
\midrule
& CLIP    
& 84.68                
& {\ul 86.18}    
& \textbf{87.35} \\
Pets                
& ALIGN     
& 82.56                
& \textbf{86.02} 
& {\ul 85.69}    \\
                    
& MetaCLIP  
& {\ul 87.84}          
& \textbf{89.51} 
& 86.94          \\
                    
& CLIPA
& \textbf{91.93}
& 91.06
& {\ul 91.20}    \\
\midrule
CUB                 
& CLIP    
& {\ul 51.48}          
& \textbf{53.40} 
& 50.36          \\
                    
& CLIPA     
& {\ul 74.56}          
& 73.33          
& \textbf{76.42} \\
                    
\midrule
& CLIP    
& {\ul 59.59}          
& 55.79          
& \textbf{60.56} \\
                    
& ALIGN     
& 52.38                
& \textbf{55.12} 
& {\ul 53.57}    \\
Stanford
& EVA       
& {\ul 76.08}          
& 74.37          
& \textbf{78.66} \\

Dogs                    
& MetaCLIP
& \textbf{65.50}       
& 62.73          
& 65.20          \\
                    
& CLIPA     
& 77.19                
& \textbf{79.07} 
& {\ul 78.25}    \\
\bottomrule
\end{tabular}
}
\caption{Classification accuracy of descriptions that explain visual/non-visual characteristics of the concepts. First and second best numbers are in bold and underline.}
\label{tab:visual_classification}
\end{table}

\textbf{Non-visual descriptions contribute significantly to VLM representations.}
For the 19 cases in which informative descriptions alone lead to a considerable improvement, we report the percentage of informative descriptions that contain visual attributes in~\cref{tab:is_visual}.
In four cases, non-visual descriptions are almost solely responsible for the accuracy boosts, and only a small fraction (<25\%) of informative descriptions contain visual information.
To study the role of non-visual information in the remaining cases, we split the informative descriptions into visual and non-visual categories and measure the classification accuracy (\cref{tab:visual_classification}).
Although visual information contributes to VLM representations, in 11 out of 15 cases, non-visual information alone considerably boosts accuracy.
In total, in 15 out of 19 cases, non-visual information is either the dominant factor or contributes significantly to VLM representations.
Even more concerning, in only four of the 15 cases, descriptions with visual information perform better than the ones with non-visual information by a considerable margin (>1\% accuracy boost).

\subsection{Described Attributes Across Different VLMs and Datasets}

\begin{figure*}[t]
\centering

\begin{subfigure}{\textwidth}
\centering
\begin{subfigure}{77mm}
\centering
\includegraphics[width=\textwidth]{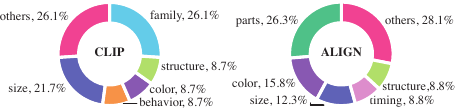}
\caption{Flowers}
\end{subfigure}
\hspace{4mm}
\begin{subfigure}{77mm}
\centering
\includegraphics[width=\textwidth]{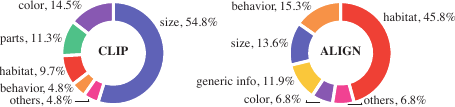} 
\caption{CUB}
\end{subfigure}
\end{subfigure}

\caption{Most common described attributes for CLIP and ALIGN for CUB and Flowers. Different VLMs prioritize different attributes to represent concepts. Even the same VLM prioritizes different attributes across datasets.}
\label{fig:pie_charts_main}
\end{figure*}

To understand how the described attributes change for different datasets and VLMs, we manually extract and list the attributes for 50 randomly selected descriptions for CLIP and ALIGN models on Flowers and CUB datasets (200 in total).
\Cref{fig:pie_charts_main} shows the most common attributes in each case.
Note that each description might include multiple attributes.

\textbf{Different VLMs prioritize different attributes.}
We observe that even for the same dataset, different VLMs represent concepts differently.
For instance, CLIP relies more on ``family'' and ``size'' attributes to represent flowers, while ALIGN relies more on ``parts'' and ``color'' attributes (\cref{fig:pie_charts_main}).
To verify that different VLMs represent concepts differently, we use the preferred descriptions of one VLM for classification with the other VLM.
Each VLM performs better with descriptions that include its preferred attributes, i.e., attributes that contribute more to its concept representations, confirming the difference in representations across VLMs (\cref{tab:cross_vlm}).

\begin{table}[h]
\centering
\begin{tabular}{llcc}
\toprule

Dataset
& VLM
& CLIP
& ALIGN
\\
\midrule

\multirow{2}{*}{Flowers}
& CLIP
& \textbf{73.65}
& 69.82
\\

& ALIGN
& 61.03
& \textbf{62.19}
\\
\midrule

\multirow{2}{*}{CUB}
& CLIP
& \textbf{53.54}
& 51.64
\\

& ALIGN 
& 37.45
& \textbf{39.66}
\\
\bottomrule
\end{tabular}
\caption{Cross VLM accuracy on Flowers and CUB datasets. Rows and columns represent VLMs used for classification and reward calculation, respectively.}
\label{tab:cross_vlm}
\end{table}

\textbf{VLMs prioritize different attributes for different datasets.}
The same VLM prioritizes different attributes for representing bird and flower species (\cref{fig:pie_charts_main}).
These results suggest that VLM's behavior should be studied on various datasets to draw conclusions about their capabilities.
EX2 is well suited to automatically carry out such extensive analysis on a wide range of classification datasets.

\subsection{Applications for Hypothesis Generation}
\label{sec:hypothesis_gen}
The trends in EX2 descriptions provide helpful cues for discovering new research questions.
For instance, we notice that SigLIP descriptions often resemble software instructions or photo credits (e.g., \rinlinecode{Click to enlarge image of CONCEPT}).
Although WebLi, SigLIP's pre-training dataset, is not publicly available, we know that many of the image-text pairs are created with OCR~\citep{chen2023pali}, and perhaps they are more likely to resemble such specific patterns.
Thus, one potential hypothesis is that OCR, as a data source, could lead to more model bias towards spurious patterns.
In a preliminary experiment, we engineer three prompt templates that mimic these patterns and use them for classification with CLIP and SigLIP on Stanford Dogs.
The custom templates improve the accuracy for SigLIP but not for CLIP compared to the commonly used generic template (\cref{tab:custom_temp}).
Our preliminary results encourage future work to systematically study the impact of such patterns on SigLIP and other VLMs and investigate the role of the pre-training dataset in such behaviors, which could provide valuable insights for creating the next generation of image-text datasets.

\begin{table}[h]
\centering
\begin{tabular}{p{0.5\columnwidth}cc}
\toprule

Description
& CLIP
& SigLIP
\\
\midrule

\footnotesize
\tt
A photo of a \{class name\}.
&
58.52
&
77.23
\\

\midrule

\footnotesize
\tt
A photo of a \{class name\}. Click to enlarge.
&
58.26
&
78.24
\\
\midrule
                               
\footnotesize
\tt
Below is a picture of a \{class name\}. I obtained this photo from the following site
&
58.48
&
77.75
\\
\midrule
                               
\footnotesize
\tt
Click here to view the photos of a \{class name\}.
&
58.46
&
77.94
\\
\bottomrule
\end{tabular}
\caption{The accuracy of CLIP and SigLIP on Stanford Dogs dataset with custom prompt templates.}
\label{tab:custom_temp}
\end{table}

\section{Additional Analysis}
\label{sec:additional_analysis}

\noindent\textbf{ChatGPT Accuracy}\hphantom{A}
To assess the accuracy of ChatGPT~\citep{brown2020language,ouyang2022training} inspections, we randomly select 50 descriptions from each of the VLM-dataset combinations (2100 in total) and manually check if they are informative and contain visual attributes.
Then, we compare the results to ChatGPT responses.
ChatGPT correctly classifies the descriptions most of the time and agrees with our judgment 86.3\% and 88.1\% of the time for checking if descriptions are informative and contain visual attributes, respectively (\cref{tab:gpt_agreement} in \cref{sec:appendix_gpt_agreement}).

\noindent\textbf{KL Penalty}\hphantom{A}
Since the KL penalty in the reward function influences both the accuracy and amount of helpful information in descriptions, here we study its impact on our findings.
We repeat our experiments for CUB and Flowers datasets with a reduced KL penalty.
Since we use the adaptive KL coefficient technique of~\citet{ziegler2019fine} to set the trade-off between VLM preferences and KL divergence, we increase the target KL value in the adaptive KL coefficient technique from 10 in the main experiments to 20 for this experiment.

As reported in~\cref{tab:diff_kl} in \cref{sec:appendix_kl_analysis}, in most cases, reducing the KL penalty (i.e., more emphasis on VLM preferences) decreases the amount of helpful information in descriptions, which further strengthens our findings about the major role of spurious descriptions in VLM representations.

\section{Qualitative Results}

\begin{table*}[t]
\small
\centering
\begin{tabular}{p{0.05\textwidth}p{0.19\textwidth}p{0.19\textwidth}p{0.19\textwidth}p{0.19\textwidth}}
\toprule

Query
&
\tt
\scriptsize
Describe a photo of a Yellow-billed Cuckoo.
&
\tt
\scriptsize
Write a story or narrative inspired by a photo of a Acadian Flycatcher.
&
\tt
\scriptsize
How does a photo of a grape hyacinth look like?
&
\tt
\scriptsize
How does a photo of a lenten rose look like?
\\
\midrule

\hphantom{\scriptsize{A}}
CLIP
&
\tt
\scriptsize
The Yellow-billed cuckool is a medium-sized bird with a dark brown to blackish brown plumage. The average length of a
&
\tt
\scriptsize
The Acadian Flycatcher is a medium sized bird, a common spring and summer migrant in eastern North America. Acadian Flycatchers
&
\tt
\scriptsize
A grape hyacinth plant (Muscari sp.) has two bright blue flowers at its tip per stalk. The flowers have small, colored dots
&
\tt
\scriptsize
"\#\# How does a lenten Rose look?

Lenten rose (Helleborus spp.) is a perennial plant native to Europe"
\\
\midrule

\hphantom{\scriptsize{A}}
SigLIP
&
\tt
\scriptsize
"Photo of a Yellow-billed Cuckoo. This image was downloaded from the US Fish \& Wildlife Service website. Here's a"
&
\tt
\scriptsize
"It's not until after I direct your attention to it that you see the Acadian Flycatcher.

Before I point it out,"
&
\tt
\scriptsize
"Here is a grape hyacinth photo that I took.

Grape hyacinths look like quite elegant, pretty and delicate flowers,"
&
\tt
\scriptsize
"A lenten rose is among my favorite flowers. This post will show you how a photo of a lenten rose looks like. A photo"
\\

\bottomrule
\end{tabular}
\caption{Aligned descriptions generated in response to four different queries for CLIP and SigLIP.}
\label{tab:samples_main_body}
\end{table*}

\begin{table*}[h]
\small
\centering
\begin{tabular}{p{0.4\textwidth}p{0.4\textwidth}c}
\toprule

CLIP
& SigLIP
& \makecell[c]{\# Batches\\Seen}
\\
\midrule

\scriptsize
\tt
The Chihuahua is a Mexican dog breed, however, this breed is not the smallest. The small size is a small – sized breed,
&
\scriptsize
\tt
A Chihuahua is a Mexican dog breed known for its small size, large ears, and wrinkled skin. It is believed to be origin 
&
20
\\
\midrule

\scriptsize
\tt
\#\# What is a Chihuahua dog?

Chihuahua is a small dog that was originated in the Chihuah
&
\scriptsize
\tt
Chihuahuas are small dogs with a big personality. The most popular dog breed, Chihuahuas are tireless, gentle and loyal. 
&
200
\\
\midrule

\scriptsize
\tt
\#\# Describe a photo of a Chihuahua dog

A Chihuahua is a small dog that originated in the Mexican state of
&
\scriptsize
\tt
Chihuahuas are my favorite kind of dog, however, this one is not the most lovely. He is sure of himself – and that's 
&
400
\\
\midrule

\scriptsize
\tt
\#\# Describe a photo of a Chihuahua dog

A Chihuahua is a small dog breed, named after the Mexican state of
&
\scriptsize
\tt
Here's a photo of my Chihuahua dog, Ginger who is quite a lovely dog and is sure to win over everyone's love 
&
600
\\
\midrule

\scriptsize
\tt
\#\# Describe a photo of a Chihuahua dog

A Chihuahua is a small dog breed, named after the Mexican state of
&
\scriptsize
\tt
Here's a photo of my Chihuahua dog Huey who passed away at the age of 16 on 6/10 
&
800
\\
\midrule

\scriptsize
\tt
\#\# Describe a photo of a Chihuahua dog

A Chihuahua is a small dog with a short, smooth coat.
&
\scriptsize
\tt
Here's a photo of my Chihuahua dog Huey who passed away at eleven years old a few years ago – 
&
999
\\

\bottomrule
\end{tabular}
\caption{Examples of how descriptions change during training for CLIP and SigLIP for the same query.}
\label{tab:desc_change_main_body}
\vspace{-3mm}
\end{table*}

\Cref{tab:samples_main_body} shows the generated descriptions for CLIP and SigLIP in response to four sample queries.
Consistent with our previous findings, different VLMs prefer different types of descriptions for the same concept.
For instance, while CLIP's preferred descriptions explain the visual appearance of concepts, SigLIP's preferred descriptions are often spurious and rarely provide helpful information for classifying images.
See~\cref{sec:qual_observ} for generated descriptions for other VLMs.

Although we use the same base LLM and fine-tuning hyperparameters for all experiments, generated descriptions diverge based on VLM preferences. 
\Cref{tab:desc_change_main_body} shows how descriptions change during fine-tuning with CLIP and SigLIP rewards.
As expected, in both cases, the LLM initially describes the general characteristics of a Chihuahua dog. 
As fine-tuning progresses, descriptions generated with CLIP rewards keep explaining what a Chihuahua dog is but use different attributes throughout fine-tuning.
On the other hand, with SigLIP rewards, information helpful for identifying images of Chihuahua dogs decreases during fine-tuning, and descriptions finally converge to personal stories involving a Chihuahua dog.
See~\cref{sec:qual_observ} for more examples.

\section{Conclusion}
\label{sec:conclusion}
In this work, we introduce Extract and Explore (EX2), a novel approach for characterizing features that contribute to VLM representations.
EX2 uses RL to align an LLM with VLM preferences and generates descriptions that include features that are important for the VLM.
EX2 then examines the descriptions to identify the common features that contribute to VLM representations.
Using EX2, we show that spurious descriptions contribute significantly to VLM representations despite providing no helpful information.
VLMs also rely significantly on non-visual attributes to represent visual concepts.
Moreover, different VLMs prioritize different attributes to represent similar concepts.
Even the same VLM prioritizes different attributes across datasets.
Our results encourage future work to address the significant impact of non-visual and spurious information on VLM representations.
EX2 provides future work with a flexible tool to study various aspects of VLM representations.

\section*{Limitations}

\noindent\textbf{RL Stability}\hphantom{A}
EX2 uses RL to discover the model preferences.
Given the sensitivity of RL algorithms~\citep{engstrom2020implementation,andrychowicz2020matters}, the generated descriptions should be interpreted cautiously.
In this work, our conclusions are based on the sensitivity of VLMs to spurious and non-visual descriptions, and we use classification accuracy to verify their impact on VLMs.
Future work should consider the sensitivity of RL algorithms and use similar measures to verify their conclusions.
More broadly, considering the benefits of EX2, we encourage future work to study more robust approaches for learning model preferences for analysis purposes.

\noindent\textbf{LLM Capabilities}\hphantom{A}
Regardless of the optimization approach, to successfully align with VLM preferences, the original LLM should be capable of generating the desired descriptions (e.g., be aware of the required world knowledge).
For instance, in our early experiments, Llama2-7B~\citep{touvron2023llama} failed to learn helpful features for each VLM.
In this work, we mainly focus on high-level features (i.e., the presence of helpful and visual information), and our results show that our LLM successfully describes the target classes from various perspectives.
We encourage future work to further study the role of the LLM, especially for investigating more nuanced characteristics of the descriptions, like style or linguistic properties.

\noindent\textbf{ChatGPT as Inspector}\hphantom{A}
Although we can use handcrafted rules to inspect the low-level properties of descriptions like word frequency or number of tokens, the same is not possible for high-level characteristics (e.g., whether description contents are spurious or non-visual).
As a result, we resort to ChatGPT to inspect the large number of descriptions at scale.
We also do a manual evaluation of the quality of ChatGPT inspections and get reasonable results (\cref{sec:additional_analysis}).
However, using ChatGPT for inspections poses a limitation when it comes to more nuanced or complex characteristics of descriptions that are challenging for current LLMs to detect reliably.
One potential path for future work is to train inspector models that check for a specific characteristic of descriptions.
This also introduces new challenges like collecting training data and evaluating the customized inspector models.

\section*{Acknowledgements}
We gratefully acknowledge support from Cisco and Cognex.
Disclosure: Stephen Bach is an advisor to Snorkel AI, a company that provides software and services for data-centric artificial intelligence.
\bibliography{misc/refs_amd}

\appendix

\begin{table}[h]
\centering

\begin{tabular}{@{}lc@{}}
\toprule
Query
& Accuracy
\\
\midrule

Informative
& 86.33
\\

Visual
& 88.14                            
\\
\bottomrule
\end{tabular}
\caption{Agreement rate between ChatGPT and our manual inspection for informative and visual checks for 500 randomly selected descriptions.}
\label{tab:gpt_agreement}
\vspace{-3mm}
\end{table}

\section{Discussion}
\label{sec:discussion}

\subsection{Analysis Design and Rationale}
\label{sec:discussion_rationale}

As discussed in~\cref{sec:explore}, we consider descriptions to be spurious if they do not provide any additional information about the concepts beyond concept names.
In other words, descriptions that do not contain any of the concept attributes are spurious.
Furthermore, we consider informative descriptions to be non-visual if they do not contain any attributes related to the physical appearance of the concepts.
\cref{tab:category_examples} provides several examples of this categorization.

\begin{table*}[t]
\centering
\begin{tabular}{p{0.6\textwidth}ccc}
\toprule

Description
& Spurious
& Non-visual
& Visual
\\
\midrule

\footnotesize
\tt
This photo of a camellia is an underly realistic representation of its real-life counterpart.
&
\checkmark
&
&
\\
\midrule

\footnotesize
\tt
This is a picture of globen thistle. Your assigned task is to explain as thoroughly as you can what this picture of the globe thistle represents
&
\checkmark
&
&
\\
\midrule

\footnotesize
\tt
A primula in the morning glow of the sunlight. A photo that has instigated many a point of view from numerous people. 
&
\checkmark
&
&
\\
\midrule

\footnotesize
\tt
The Downy Woodpecker is one of several species that I see quite frequently birding in and around my home.
&
\checkmark
&
&
\\
\midrule

\footnotesize
\tt
It isn't too common to see an Indigo Bunting around our neighborhood as they generally don't spend too much time in one place, 
&
\checkmark
&
&
\\
\midrule

\footnotesize
\tt
The Laysan Albatross is a fascinating bird. I have had the privilege of seeing this bird on only a few occasions, 
&
\checkmark
&
&
\\
\midrule

\footnotesize
\tt
I found this picture of a giant white arum lily. I have personally never found an arum lily, let alone even close to this size and
&
\checkmark
&
&
\\
\midrule

\footnotesize
\tt
The bobolink, is a North American bird. Bobolinks are very common in U.S. and Near‚bobolinks, are rarely
&
&
\checkmark
&
\\
\midrule

\footnotesize
\tt
There are many varieties of hummingbirds in North America.
&
&
\checkmark
&
\\
\midrule

\footnotesize
\tt
English Cocker Spaniels are an intelligent, determined and energetic dog that excels in a number of task activities.
&
&
\checkmark
&
\\
\midrule

\footnotesize
\tt
American bulldogs are amazing dogs that are gentle, smart, and playful.
&
&
\checkmark
&
\\
\midrule

\footnotesize
\tt
A primula is a flower that can be found in the cooler temperatures of winter and Spring.
&
&
\checkmark
&
\\
\midrule

\footnotesize
\tt
The Red-headed woodpecker has black and white coloration, a tightly marked undertail and a small crest. 
&
&
&
\checkmark
\\
\midrule

\footnotesize
\tt
The Winter Wren is small, slitty, round-tailed and is covered with brown and grey eyes.
&
&
&
\checkmark
\\
\midrule

\footnotesize
\tt
A British Shorthair is round, short-haired cat with a muscular body and luxurious fuzzy fur.
&
&
&
\checkmark
\\
\midrule

\footnotesize
\tt
Bassett hounds are breed of dogs famous for big drop ears and hung short legs.
&
&
&
\checkmark
\\
\midrule

\footnotesize
\tt
Pelargoniums are a type of flowering shrub characterized by their distinctly round leaves and rather whispy stems. 
&
&
&
\checkmark
\\

\bottomrule
\end{tabular}
\caption{Examples of spurious, non-visual, and visual descriptions.}
\label{tab:category_examples}
\end{table*}

In our analysis, we use a specific definition of spurious and non-visual descriptions and argue that sensitivity to these types of descriptions is undesirable (\cref{sec:explore}).
Although this is not universally applicable, VLMs that do not take into account spurious and non-visual information are desirable for many applications.
For example, we can efficiently adapt such VLMs to new classification tasks by just describing the concepts’ visual appearance to achieve good performance, eliminating the need for expensive prompt tuning or prompt engineering to find model-specific prompts that boost the performance.
Another example is more reliable deployment in practice.
Ideally, we want the same behavior in response to user instructions (e.g. pour the coffee beans in the bag) regardless of the variation in users’ phrasing and wording (i.e., instances of spurious correlations) or the presence of non-visual information (e.g., the smell of coffee beans).

We emphasize that the definition of characteristics and whether they are desirable is decided by the user and not restricted by EX2.

\subsection{Concept-level Analysis}
\label{sec:discussion_concept_level}

In this work, we analyze VLMs at the concept level rather than the image level.
We calculate our reward score for each concept rather than a single image, i.e., the reward is calculated over a set of 250 images for each concept.
Similarly, our definition of spurious and non-visual descriptions focuses on the concept in the images and not the entire scene (e.g., the background).
The insights from our concept-level analysis are important for many applications, such as the ones mentioned in~\cref{sec:discussion_rationale}.
We encourage future work to also conduct an image-level analysis of VLMs.
Such analysis is especially important to understand how VLMs perceive more complex scenes, like the relation between multiple objects in the image~\citep{lewis2022does}.

\subsection{Correlation between Findings and Training Details}
\label{sec:discussion_findings_cause}

Our observations suggest a correlation between training data and our findings.
For instance, CLIP, MetaCLIP, and DFN have similar (but not identical) training setups, except for the training data, where MetaCLIP attempts to recreate CLIP’s pre-training data.
As a result, we observe that MetaCLIP prioritizes informative descriptions more often than DFN but less often than CLIP (\cref{tab:has_info}).
SigLIP is another interesting case.
Unlike other VLMs, SigLIP’s training dataset, WebLI, uses OCR results as text descriptions.
We observe that compared to other VLMs, SigLIP benefits more often from spurious descriptions.
Also in~\cref{sec:hypothesis_gen}, we show that SigLIP is biased towards spurious patterns that are expected in OCR data.

We do not make any claims about correlation between specific properties of training data and our findings.
Such a conclusion requires extensive and systematic experiments that train multiple variations of the same VLM and control for other training details.
We believe EX2 is well-suited to provide insights and generate new hypotheses that guide such extensive analysis and hope it helps future work to study the impacts of training data on VLMs.

\section{Qualitative Observations}
\label{sec:qual_observ}

\textbf{Descriptions for different VLMs.}\hphantom{A}
\Cref{tab:samples_main} shows the generated descriptions for different VLMs in response to four sample queries.
Consistent with our findings, we observe that different VLMs prefer different descriptions for the same concept.

\textbf{Name-only descriptions.}\hphantom{A}
For several VLM-dataset combinations, the LLM learns to rely more on concept names rather than coherent English sentences.
In these cases, the ratio of informative descriptions is very small (\cref{tab:has_info}). 
In~\cref{tab:name_only_descs}, we provide examples of such descriptions for VLM-dataset combinations with less than 10\% informative descriptions in~\cref{tab:has_info}.

\begin{table*}[t]
\centering
\begin{tabular}{ccp{0.7\textwidth}}

\toprule

VLM
& Dataset
&
Description
\\
\midrule

\multirow{4}{*}{EVA}
&
\multirow{2}{*}{Flowers}
&
\footnotesize
\tt
Pink-yellow dahlia, dahlia, dahlia, dahlia, dahlia, dahlia, dahlia
\\
\cmidrule(l){3-3}

&

&
\footnotesize
\tt
\#\# thorn apple

\#\#\# thorn apple

\#\#\#\# thorn apple

\#\#\#\#\# thorn apple

\#\#\#\#\#\# thorn apple
\\
\cmidrule(l){2-3}

&
\multirow{2}{*}{Pets}
&
\footnotesize
\tt
A Havanese
\\
\cmidrule(l){3-3}

&
&
\footnotesize
\tt
A keeshond
\\
\midrule

\multirow{4}{*}{SigLIP}
&
\multirow{2}{*}{Aircrafts}
&
\footnotesize
\tt
McDonnell Douglas DC-9-30 airplane

McDonnell Douglas DC-9-30 aircraft, Photos, Hugh 2
\\
\cmidrule(l){3-3}

&

&
\footnotesize
\tt
\#\# Boeing 747-100 Aircraft Specifications

The Boeing 747-100(650
\\
\cmidrule(l){2-3}

&
\multirow{2}{*}{Cars}
&
\footnotesize
\tt
\#\#\# A Photo of a 2012 Toyota 4Runner SUV

The image shows a 2012 Toyota 4
\\
\cmidrule(l){3-3}

&
&
\footnotesize
\tt
\#\# A photo of a 2007 Chevrolet Monte Carlo Coupe

A photo of a 2007 Chevrolet
\\
\midrule

\multirow{2}{*}{CLIPA}
&
\multirow{2}{*}{Aircrafts}
&
\footnotesize
\tt
Embraer EMB-120

The Embraer EMB-120

The Embraer EMB-12
\\
\cmidrule(l){3-3}

&

&
\footnotesize
\tt
\#\# McDonnell Douglas MD-87

McDonnell Douglas MD-87. Credits: Wikipedia

The McDon
\\
\midrule

\multirow{2}{*}{DFN}
&
\multirow{2}{*}{CUB}
&
\footnotesize
\tt
Canada Warbler Photo 1

Canada Warbler Photo 2

Canada Warbler Photo 3

Canada Warbler Photo 4
\\
\cmidrule(l){3-3}

&

&
\footnotesize
\tt
Worm-eating Warbler 1

Worm-eating Warbler 2

Worm-eating Warbler 3
\\

\bottomrule
\end{tabular}
\caption{Examples of name-only descriptions generated by the LLM.}
\label{tab:name_only_descs}
\end{table*}

\textbf{Descriptions throughout training.}\hphantom{A}
Although we use the same LLM to learn the preferences of all VLMs, the LLM adapts according to VLM preferences throughout training.
In~\cref{tab:desc_change_1,tab:desc_change_2}, we show how LLM descriptions change during training for two different VLMs in response to the same query.

\begin{table*}[t]
\centering
\begin{tabular}{p{0.4\textwidth}p{0.4\textwidth}c}
\toprule

CLIP
& SigLIP
& \makecell[c]{\# batches\\seen}
\\
\midrule

\footnotesize
\tt
The Chihuahua is a Mexican dog breed, however, this breed is not the smallest. The small size is a small – sized breed,
&
\footnotesize
\tt
A Chihuahua is a Mexican dog breed known for its small size, large ears, and wrinkled skin. It is believed to be origin 
&
20
\\
\midrule

\footnotesize
\tt
\#\# What is a Chihuahua dog?

Chihuahua is a small dog that was bred in the United Kingdom.
&
\footnotesize
\tt
A Chihuahua´s eyes are wide, curious and lively. Sometimes the head of a small Chihuahua is flat.
&
100
\\
\midrule

\footnotesize
\tt
\#\# What is a Chihuahua dog?

Chihuahua is a small dog that was originated in the Chihuah
&
\footnotesize
\tt
Chihuahuas are small dogs with a big personality. The most popular dog breed, Chihuahuas are tireless, gentle and loyal. 
&
200
\\
\midrule

\footnotesize
\tt
\#\# What Is a Chihuahua Dog?

- A Chihuahua dog is small, thin, and cute.
- They
&
\footnotesize
\tt
Chihuahuas are tiny dogs with a big spirit. The most popular dog breed, Chihuahuas are small, but have big personalities 
&
300
\\
\midrule

\footnotesize
\tt
\#\# Describe a photo of a Chihuahua dog

A Chihuahua is a small dog that originated in the Mexican state of
&
\footnotesize
\tt
Chihuahuas are my favorite kind of dog, however, this one is not the most lovely. He is sure of himself – and that's 
&
400
\\
\midrule

\footnotesize
\tt
\#\# Describe a photo of a Chihuahua dog

A Chihuahua is a small dog breed which originates from the state of
&
\footnotesize
\tt
Chihuahuas are tiny dogs! Look at them on the photo!

If you don't know what a Chihuahua dog
&
500
\\
\midrule

\footnotesize
\tt
\#\# Describe a photo of a Chihuahua dog

A Chihuahua is a small dog breed, named after the Mexican state of
&
\footnotesize
\tt
Here's a photo of my Chihuahua dog, Ginger who is quite a lovely dog and is sure to win over everyone's love 
&
600
\\
\midrule

\footnotesize
\tt
\#\# Describe a photo of a Chihuahua dog

A Chihuahua is a small dog. According to the American Kennel Club
&
\footnotesize
\tt
Here's a photo of my Chihuahua dog, Ginger who is quite a lovely dog and is sure to win over everyone's love 
&
700
\\
\midrule

\footnotesize
\tt
\#\# Describe a photo of a Chihuahua dog

A Chihuahua is a small dog breed, named after the Mexican state of
&
\footnotesize
\tt
Here's a photo of my Chihuahua dog Huey who passed away at the age of 16 on 6/10 
&
800
\\
\midrule

\footnotesize
\tt
\#\# Describe a photo of a Chihuahua dog

Chihuahua dogs are among the best-known dogs in the world.
&
\footnotesize
\tt
Here's a photo of Rachel's Chihuahua. I chose this photo because there is older one but this one is of particular Chih 
&
900
\\
\midrule

\footnotesize
\tt
\#\# Describe a photo of a Chihuahua dog

A Chihuahua is a small dog with a short, smooth coat. There
&
\footnotesize
\tt
Here's a photo of my Chihuahua dog Huey who passed away at eleven years old a few years ago – 
&
999
\\

\bottomrule
\end{tabular}
\caption{Examples of how descriptions change during training for CLIP and SigLIP for the same query. For CLIP, the LLM attempts to describe a Chihuahua dog in all steps but the type of information in descriptions change during training. However, for SigLIP, the LLM first starts by describing a Chihuahua dog but then converges to generating personal stories about Chihuahua dog as a pet.}
\label{tab:desc_change_1}
\end{table*}

\begin{table*}[t]
\centering
\begin{tabular}{p{0.4\textwidth}p{0.4\textwidth}c}
\toprule

CLIP
& SigLIP
& \makecell[c]{\# batches\\seen}
\\
\midrule

\footnotesize
\tt
The African hunting dog, also known as the African wild dog or Painted Dog, is a graceful and beautiful breed. The African hunting dog is a type
&
\footnotesize
\tt
The african hunting dog (Lycaon pictus), also known as the painted dog, is a medium-sized hunting canine native to Africa
&
20
\\
\midrule

\footnotesize
\tt
\#\# The African Hunting Dog – A Look at the

The African hunting dog is a breed of dog that is known for its hunting abilities. The African
&
\footnotesize
\tt
The African hunting dog is the tallest breed of dog. It has a tufted tail and a white coat that may have patterns, and its eyes may
&
100
\\
\midrule

\footnotesize
\tt
\#\# The African Hunting Dog – A Look at the

The African hunting dog is a unique and interesting breed of dog that is native to Africa. They
&
\footnotesize
\tt
1. If you don’t have a image of an african hunting dog you can use one from the internet (e.g., shapingdog
&
200
\\
\midrule

\footnotesize
\tt
\#\# The African Hunting Dog – A Look at the Hunting Dogs of Africa

The African hunting dog is a type of hunting dog. They are
&
\footnotesize
\tt
The African hunting dog (Lycaon pictus) is not very long in length, but it is rather large in weight. As strong as woods are
&
300
\\
\midrule

\footnotesize
\tt
\#\# Introduction to African Hunting Dogs

African Hunting Dogs are one of the most interesting and also adorable dog breeds. They
&
\footnotesize
\tt
The African hunting dog (Lycaon pictus) is not very long in length, but it is rather large in stature. The hound stands
&
400
\\
\midrule

\footnotesize
\tt
1. Introduction of the African hunting dog

The African hunting dog is a wild canid that lives in Africa and also adapts perfectly to different types
&
\footnotesize
\tt
Of the photos of African hunting dogs that I have found, the photo below stood out to me. It is a photo of an African hunting dog in the field
&
500
\\
\midrule

\footnotesize
\tt
1. Introduction of the African hunting dog:

The African hunting dog is a canid that lives in Africa. It is known for its shaggy
&
\footnotesize
\tt
The picture that I chose pictures an African hunting dog and the first thing that I noticed from this picture was how the belly of the dog was hanging out while the
&
600
\\
\midrule

\footnotesize
\tt
1. Introduction of the African hunting dog

The African hunting dog is a medium-sized dog that weighs around 10-30 pounds and
&
\footnotesize
\tt
I found a photograph of an African Hunting Dog on Google I thought was half good and half bad. I was trying to find a photograph that didn't
&
700
\\
\midrule

\footnotesize
\tt
1. Introduction of the African hunting dog

The African hunting dog is a wild canid that lives in Africa and also adapts perfectly to different types
&
\footnotesize
\tt
This photograph features a African hunting dog. The composition in the photograph is lacking as it seems somewhat chaotic. I do, however, quite like the quick crop
&
800
\\
\midrule

\footnotesize
\tt
\#\# Introduction to African Hunting Dogs

African Hunting Dogs are one of the most interesting and also adorable animals on the continent.
&
\footnotesize
\tt
Today we will be critiqueing our African hunting dog. Below is a link to a photograph of the dog.

African hunting dog.
&
900
\\
\midrule

\footnotesize
\tt
\#\# The African Hunting Dog – A Perfect Motion

The African Dog is a very unique animal. It is known for its ability to run at very
&
\footnotesize
\tt
Here is an example photograph of an African hunting dog.

Click here to see visualization as it needs to as well.

Permission and credits:
&
999
\\

\bottomrule
\end{tabular}
\caption{Examples of how descriptions change during training for CLIP and SigLIP for the same query. For CLIP, the LLM attempts to describe an African hunting dog in most steps but the type of information in descriptions change during training. However, for SigLIP, the LLM first starts by describing an African hunting dog but then converges to generating spurious information like the download link for an image of an African hunting dog.}
\label{tab:desc_change_2}
\end{table*}

\section{Implementation Details}
\label{sec:imp_details}

We use the same hyperparameters as~\citet{ziegler2019fine} used for style tasks, which are also the default hyperparameters in TRL library~\footnote{\url{https://huggingface.co/docs/trl/index}} that we use in our experiments.
We change the batch size to 32 and accumulate the gradients every eight steps.
For the adaptive KL algorithm, we use a target KL of 10.
To reduce computational requirements, we only use 256 images per class to calculate the reward for each description during training.
To be able to train large models on a single GPU, we use 8-bit Q-LORA with rank $r=16$~\citep{dettmers2024qlora}.
Each experiment requires about 36 GPU hours for training on a single A6000 card with 48GB of memory.

\subsection{Prompting Details}
\label{sec:prompt_details}

To search a diverse set of descriptions, we use the 25 question templates in~\cref{tab:question_temps} to query the LLM to describe various aspects of the target concepts.

We use ChatGPT (\rinlinecode{gpt-3.5-turbo-1106}~\footnote{\url{https://openai.com/blog/openai-api}}) to inspect the large number of descriptions.
We use the prompt templates in~\cref{tab:info_check_template} and~\cref{tab:is_visual_template} to check if each description is informative and if it explains the visual attributes of the target concept, respectively.

\begin{table*}[t]
\centering
\begin{tabular}{@{}lccccccccccc@{}}
\toprule

& \multicolumn{5}{c}{Flowers}
&
& \multicolumn{5}{c}{CUB}
\\
\cmidrule(lr){2-12}

& \multicolumn{2}{c}{Original KL}
&
& \multicolumn{2}{c}{Relaxed KL}
&
& \multicolumn{2}{c}{Original KL}
&
& \multicolumn{2}{c}{Relaxed KL}
\\
\cmidrule(lr){2-6}
\cmidrule(lr){8-12}
 
& Acc
& w/ Info
&
& Acc
& w/ Info
&
& Acc
& w/ Info
&
& Acc
& w/ Info
\\
\cmidrule(lr){2-3}
\cmidrule(lr){5-6}
\cmidrule(lr){8-9}
\cmidrule(lr){11-12}
 
CLIP     
&  73.65       
&  55.33          
&
&  68.92     
&  44.08       
&
&  53.54       
&  83.50          
&
&  53.75
&  90.04       
\\

ALIGN    
&  62.19       
&  63.92          
&
&  63.99     
&  60.47       
&
&  39.66       
&  63.66          
&
&  39.68
&  67.88       
\\

EVA      
&  79.05       
&  1.80           
&
&  78.68     
&  0.16        
&
&  73.52       
&  0.04           
&
&  75.35
&  0.12        
\\

SigLIP   
&  86.81       
&  44.59          
&
&  86.45     
&  41.76       
&
&  65.33       
&  23.06          
&
&  67.09
&  9.90        
\\

MetaCLIP 
&  73.18       
&  43.06          
&
&  74.53     
&  17.73       
&
&  69.99       
&  26.58          
&
&  68.43
&  0.20        
\\

CLIPA    
&  79.61       
&  55.18          
&
&  81.20     
&  61.61       
&
&  76.99       
&  47.84          
&
&  77.10
&  36.78       
\\

DFN      
&  88.86       
&  11.69
&
&  89.06     
&  3.73        
&
&  87.78       
&  0.32           
&
&  87.92
&  0.00        
\\
\bottomrule

\end{tabular}
\caption{Classification accuracy and percentage of informative descriptions for Flowers and CUB with relaxed KL penalty (target KL of 20 instead of the original 10 in main experiments). In most cases, reducing the KL penalty (i.e., more emphasis on VLM preferences) decreases the number of informative descriptions, which further strengthens our conclusions about the significant role of spurious descriptions in VLM representations.}
\label{tab:diff_kl}
\end{table*}

\section{Models}
\label{sec:models}

We use Mistral-7B as our LLM~\citep{jiang2023mistral}.
We choose seven different VLMs for our analysis.
In parentheses, we mention the vision transformer backbone for each model~\citep{dosovitskiy2021an}.
CLIP is trained with a contrastive loss on a private dataset of 400M image-text pairs~\citep{radford2021learning} (ViT-B-32).
ALIGN is also trained with a contrastive loss~\citep{jia2021scaling}, and we use a checkpoint trained on CoYo~\citep{coyo700m} (base~\footnote{\url{https://huggingface.co/kakaobrain/align-base}}).
EVA~\citep{fang2023eva} is trained on Laion400M~\citep{schuhmann2022laion} to reconstruct the CLIP features (ViT-g-14).
SigLIP~\citep{zhai2023sigmoid} is a CLIP model that uses a sigmoid loss instead of softmax and is trained on WebLi dataset~\citep{chen2023pali} (ViT-B-16).
MetaCLIP attempts to recreate the CLIP pre-training dataset and uses it to train a similar model~\citep{xu2023demystifying} (ViT-B-16).
CLIPA is a CLIP model that uses input masking to improve efficiency~\citep{li2023inverse} and is trained on the Laion400M dataset (ViT-H-14-336).
DFN~\citep{fang2023data} uses a data filtering network to filter DataComp1B~\citep{gadre2023datacomp} and trains a CLIP model on the resulting dataset (ViT-H-14-378).

\section{KL Penalty Analysis}
\label{sec:appendix_kl_analysis}
To understand the impact of the KL penalty in the reward function on the aligned descriptions, we repeat our main experiments for CUB and Flowers datasets but relax the KL penalty.
As reported in~\cref{tab:diff_kl}, reducing the KL penalty (i.e., more emphasis on VLM preferences) decreases the amount of helpful information in descriptions, which further strengthens our findings about the significant role of spurious information in VLM representations.

\section{ChatGPT Accuracy}
\label{sec:appendix_gpt_agreement}

In~\cref{tab:gpt_agreement}, we report the rate of agreement between ChatGPT and our manual inspection for informative and visual description checks.

\begin{table*}[t]
\small
\centering
\begin{tabular}{|l|p{0.19\linewidth}|p{0.19\linewidth}|p{0.19\linewidth}|p{0.19\linewidth}|}
\hline

Query
&
\tt
\scriptsize
Describe a photo of a Yellow-billed Cuckoo.
&
\tt
\scriptsize
Write a story or narrative inspired by a photo of a Acadian Flycatcher.
&
\tt
\scriptsize
How does a photo of a grape hyacinth look like?
&
\tt
\scriptsize
How does a photo of a lenten rose look like?
\\
\hline

CLIP
&
\tt
\scriptsize
The Yellow-billed cuckool is a medium-sized bird with a dark brown to blackish brown plumage. The average length of a
&
\tt
\scriptsize
The Acadian Flycatcher is a medium sized bird, a common spring and summer migrant in eastern North America. Acadian Flycatchers
&
\tt
\scriptsize
A grape hyacinth plant (Muscari sp.) has two bright blue flowers at its tip per stalk. The flowers have small, colored dots
&
\tt
\scriptsize
"\#\# How does a lenten Rose look?

Lenten rose (Helleborus spp.) is a perennial plant native to Europe"
\\
\hline

ALIGN
&
\tt
\scriptsize
The Yellow-billed cuckoo is a common bird during the spring and summer season. They live throughout the Eastern and Middle United States, diving primarily
&
\tt
\scriptsize
"The Acadian Flycatcher is a common summer resident of the forested areas of central and eastern Canada.

Before I saw it in person,"
&
\tt
\scriptsize
A grape hyacinth is a beautiful flower that grows in the early spring. It has a bell-shaped flower that is blue or purple colored.
&
\tt
\scriptsize
A lenten rose flower is an unusual flower that looks like a butterfly when closed, and like a wonder flower when fully open, you can see its
\\
\hline

EVA
&
\tt
\scriptsize
Yellow-billed Cuckoo, Yellow-billed Cuckoo, Yellow-billed Cuckoo, Yellow-billed Cuckoo
&
\tt
\scriptsize
Acadian Flycatcher, Acadian Flycatcher, Acadian Flycatcher, Acadian Flycatcher, Acadian Fly
&
\tt
\scriptsize
\#\#\# Grape hyacinth, grape hyacinth, grape hyacinth, grape hyacinth, grape hyacinth
&
\tt
\scriptsize
\#\# Lenten Rose, Lenten Rose, Lenten Rose, Lenten Rose, Lenten Rose, Lenten Rose
\\
\hline

SigLIP
&
\tt
\scriptsize
"Photo of a Yellow-billed Cuckoo. This image was downloaded from the US Fish \& Wildlife Service website.

Here's a"
&
\tt
\scriptsize
"It's not until after I direct your attention to it that you see the Acadian Flycatcher.

Before I point it out, the"
&
\tt
\scriptsize
"Here is a grape hyacinth photo that I took.

Grape hyacinths look like quite elegant, pretty and delicate flowers,"
&
\tt
\scriptsize
"A lenten rose is among my favorite flowers. This post will show you how a photo of a lenten rose looks like.

A photo"
\\
\hline

MetaCLIP
&
\tt
\scriptsize
"Yellow-billed cuckoo

The Yellow-billed cuckoo is next on the list of birds that has captured my attention and become"
&
\tt
\scriptsize
It's funny how an Acadian Flycatcher is like a blur. I've seen one, but you really don't notice it so
&
\tt
\scriptsize
"A grape hyacinth is a beautiful perennial plant that can be grown indoors or outdoors.

The common photocopy of the"
&
\tt
\scriptsize
Did you know that lenten Rose is one of the first flowers to bloom in spring? Just take a look at this photo, you can see its
\\
\hline

CLIPA
&
\tt
\scriptsize
The Yellow-billed cuckoo is a distinctive bird that can be found in the forested regions of the Southwest. Even though it's a
&
\tt
\scriptsize
The Acadian Flycatcher is a small bird that is gray above with a white belly. It has a gray bill and black eye ring. However,
&
\tt
\scriptsize
A photo of a grape hyacinth. The grape hyacinth is a bulb plant perennial with a height of between 6 and
&
\tt
\scriptsize
A lenten rose flower is an unusual flower that doesn't look at all like a rose. It is short, bell-shaped, and looks like
\\
\hline

DFN
&
\tt
\scriptsize
"Yellow-billed Cuckoo Photos
Yellow-billed Cuckoo Photos
Yellow-billed Cuckoo Photos"
&
\tt
\scriptsize
"Acadian Flycatcher

Acadian Flycatcher

Acadian Flycatcher

Acadian Flycatcher
"
&
\tt
\scriptsize
"A photo of a grape hyacinth.

\# How does a photo of a grape hyacinth look like?

How does the"
&
\tt
\scriptsize
"Did you know that lenten Rose is one of the most common plants during springtime?
I'm sure you have seen lenten rose flowers"
\\
\hline
\end{tabular}
\caption{Aligned descriptions generated in response to four different queries for various VLMs.}
\label{tab:samples_main}
\end{table*}
\begin{table*}[t]
\small
\centering
\begin{tabular}{|l|p{0.8\linewidth}|}

\hline

Type & Content \\
\hline

System &
\tt You are a knowledgeable teacher. Answer the questions in JSON format.\\
\hline

User &
\tt You want to explain what a tench is to your students. Does the following text snippet mention any specific details about tench that increases your students' knowledge about tench? Answer yes or no. Provide an explanation for your answer.\newline
\newline
Text snippet: A photo of a tench, with dark green color.
\\
\hline 

Assistant&
\tt \{\newline
"explanation": "It teaches the students about the color of a tench.",\newline
"increases\_knowledge": "Yes"\newline
\}\\
\hline

User
&
\tt You want to explain what a \{Class Name\} is to your students. Does the following text snippet mention any specific details about \{Class Name\} that increases your students' knowledge about \{Class Name\}? Answer yes or no. Provide an explanation for your answer.\newline
\newline
Text snippet: \{Description\}
\\
\hline

\end{tabular}
\caption{Prompt template for ChatGPT to determine if a description provides additional information about the corresponding concept.}
\label{tab:info_check_template}
\end{table*}
\begin{table*}[t]
\small
\centering
\begin{tabular}{|l|p{0.8\linewidth}|}

\hline

Type & Content \\
\hline

System &
\tt You are a knowledgeable teacher. Answer the questions in JSON format.\\
\hline

User &
\tt You want to explain to your students how a tench looks like. Does the following text snippet mention any specific information that describes the physical appearance of a tench which helps your students recognize a tench when they see it? Answer yes or no. Provide an explanation for your answer.\newline
\newline
Text snippet: A photo of a tench, with dark green color.
\\
\hline 

Assistant&
\tt \{\newline
"explanation": "It describes the color, which helps recognize a tench.",\newline
"provides\_visual\_description": "Yes"\newline
\}\\
\hline

User
&
\tt You want to explain to your students how a \{Class Name\} looks like. Does the following text snippet mention any specific information that describes the physical appearance of a \{Class Name\} which helps your students recognize a \{Class Name\} when they see it? Answer yes or no. Provide an explanation for your answer.\newline
\newline
Text snippet: \{Description\}
\\
\hline

\end{tabular}
\caption{Prompt template for ChatGPT to determine if a description explains the physical appearance of the corresponding concept.}
\label{tab:is_visual_template}
\end{table*}
\begin{table*}[t]
\centering
\resizebox{\textwidth}{!}{
\begin{tabular}{@{}lccccc|ccccc|ccccc@{}}
\toprule
         & \multicolumn{5}{c}{Flowers}                                & \multicolumn{5}{c}{Pets}                                   & \multicolumn{5}{c}{CUB}                                    \\
         \cmidrule(lr){2-16}
         & \rotatebox[origin=c]{90}{All}  & \rotatebox[origin=c]{90}{No Info} & \rotatebox[origin=c]{90}{With Info} & \rotatebox[origin=c]{90}{Non-Visual Info} & \rotatebox[origin=c]{90}{Visual Info} & \rotatebox[origin=c]{90}{All}  & \rotatebox[origin=c]{90}{No Info} & \rotatebox[origin=c]{90}{With Info} & \rotatebox[origin=c]{90}{Non-Visual Info} & \rotatebox[origin=c]{90}{Visual Info} & \rotatebox[origin=c]{90}{All}  & \rotatebox[origin=c]{90}{No Info} & \rotatebox[origin=c]{90}{With Info} & \rotatebox[origin=c]{90}{Non-Visual Info} & \rotatebox[origin=c]{90}{Visual Info} \\
         \cmidrule(lr){2-16}
CLIP   & 2550 & 1139    & 1411      & 898             & 512         & 925  & 353     & 571       & 254             & 313         & 5000 & 825     & 4175      & 1502            & 2672        \\
ALIGN    & 2550 & 920     & 1630      & 710             & 920         & 925  & 289     & 636       & 281             & 354         & 5000 & 1817    & 3183      & 2585            & 595         \\
EVA      & 2550 & 2504    & 46        & -               & -           & 925  & 925     & 0         & -               & -           & 5000 & -    & -         & -               & -           \\
SigLIP   & 2550 & 1412    & 1137      & 369             & 767         & 925  & 752     & 173       & -               & -           & 5000 & 3846    & 1153      & -               & -           \\
MetaCLIP & 2550 & 1452    & 1098      & -               & -           & 925  & 627     & 298       & 125             & 171         & 5000 & 3670    & 1329      & -               & -           \\
CLIPA    & 2550 & 1143    & 1407      & 581             & 826         & 925  & 678     & 247       & 87              & 160         & 5000 & 2608    & 2392      & 1451            & 941         \\
DFN      & 2550 & -    & -       & -               & -           & 925  & -     & -       & -               & -           & 5000 & 4984    & 16        & -               & -           \\
\bottomrule
\toprule
         & \multicolumn{5}{c}{Stanford Dogs}                          & \multicolumn{5}{c}{Aircrafts}                              & \multicolumn{5}{c}{Stanford Cars}                          \\
         \cmidrule(lr){2-16}
         & \rotatebox[origin=c]{90}{All}  & \rotatebox[origin=c]{90}{No Info} & \rotatebox[origin=c]{90}{With Info} & \rotatebox[origin=c]{90}{Non-Visual Info} & \rotatebox[origin=c]{90}{Visual Info} & \rotatebox[origin=c]{90}{All}  & \rotatebox[origin=c]{90}{No Info} & \rotatebox[origin=c]{90}{With Info} & \rotatebox[origin=c]{90}{Non-Visual Info} & \rotatebox[origin=c]{90}{Visual Info} & \rotatebox[origin=c]{90}{All}  & \rotatebox[origin=c]{90}{No Info} & \rotatebox[origin=c]{90}{With Info} & \rotatebox[origin=c]{90}{Non-Visual Info} & \rotatebox[origin=c]{90}{Visual Info} \\
         \cmidrule(lr){2-16}
CLIP   & 3000 & 1528    & 1471      & 979             & 492         & 2500 & 1053    & 1446      & 1297            & 148         & 4900 & 2721    & 2179      & -               & -           \\
ALIGN    & 3000 & 780     & 2216      & 985             & 1231        & 2500 & 860     & 1640      & 1479            & 161         & 4900 & 2799    & 2101      & 1919            & 182         \\
EVA      & 3000 & 1902    & 1098      & 759             & 338         & 2500 & 1770    & 730       & -               & -           & 4900 & -    & -      & -            & -         \\
SigLIP   & 3000 & 2370    & 630       & -               & -           & 2500 & 2366    & 134       & -               & -           & 4900 & 4412    & 488       & -               & -           \\
MetaCLIP & 3000 & 1458    & 1542      & 841             & 700         & 2500 & -    & -        & -               & -           & 4900 & -    & -      & -               & -           \\
CLIPA    & 3000 & 633     & 2364      & 941             & 1419        & 2500 & 2344    & 156       & -               & -           & 4900 & -    & -       & -               & -           \\
DFN      & 3000 & 2312    & 687       & -               & -           & 2500 & -    & -         & -               & -           & 4900 & -    & -        & -               & -          
\\
\bottomrule
\end{tabular}
}
\caption{Breakdown of the number of informative and spurious descriptions as well visual and non-visual descriptions for experiments in which LLM-generated descriptions perform better than the generic template set.}
\label{tab:num_descs}
\end{table*}

\begin{table*}[t]
\centering
\begin{tabular}{p{0.9\textwidth}}
\toprule

\tt
Describe a photo of a CLASS NAME.
\\
\midrule

\tt
How does a photo of a CLASS NAME look like?
\\
\midrule

\tt
What are useful visual features for distinguishing a CLASS NAME in a photo?
\\
\midrule

\tt
Write a caption for an image of a CLASS NAME.
\\
\midrule

\tt
Describe the distinguishing visual characteristics of a CLASS NAME.
\\
\midrule

\tt
Describe a CLASS NAME.
\\
\midrule

\tt
How can one distinguish the images of a CLASS NAME?
\\
\midrule

\tt
Write a sentence about the visual characteristics of a CLASS NAME.
\\
\midrule

\tt
What are some unusual or creative ways to capture a CLASS NAME in a photo?
\\
\midrule

\tt
What is the typical composition or framing of a CLASS NAME in photography?
\\
\midrule

\tt
Write a poem inspired by the visual characteristics of a CLASS NAME.
\\
\midrule

\tt
Write a short story inspired by a photo of a CLASS NAME.
\\
\midrule

\tt
Critique a photo of a CLASS NAME and provide suggestions for improvement.
\\
\midrule

\tt
Interpret the symbolism in a photo of a CLASS NAME.
\\
\midrule

\tt
Compare and contrast the visual characteristics of different CLASS NAME.
\\
\midrule

\tt
What is the significance of the CLASS NAME in photography?
\\
\midrule

\tt
Identify the key elements or objects in a photo of a CLASS NAME and explain their significance.
\\
\midrule

\tt
Research and find examples of different styles or approaches to photographing a CLASS NAME and analyze their effectiveness.
\\
\midrule

\tt
Write a story or narrative inspired by a photo of a CLASS NAME.
\\
\midrule

\tt
Write a short story using a CLASS NAME as the main subject.
\\
\midrule

\tt
Analyze the composition of a photo featuring a CLASS NAME.
\\
\midrule

\tt
Write a review of a photography exhibit featuring images of CLASS NAME.
\\
\midrule

\tt
Write a critique of a photograph featuring a CLASS NAME analyzing its composition, lighting, and overall effectiveness.
\\
\midrule

\tt
Write a review of a photograph featuring a CLASS NAME and its artistic merit.
\\
\midrule

\tt
Critique a photo of a CLASS NAME and provide constructive feedback for improvement.
\\

\bottomrule
\end{tabular}
\caption{Diverse question templates to query the LLM to explain various aspects of each concept.}
\label{tab:question_temps}
\end{table*}

\end{document}